\documentclass{ifacconf}

\usepackage{graphicx}      % include this line if your document contains figures
\usepackage{natbib}        % required for bibliography
\usepackage{color}
\usepackage{subfigure}
\usepackage{amsmath}
\usepackage{amsfonts}% per il simbolo R
\usepackage{pbox}
\usepackage{soul} % per il testo barrato

\graphicspath{{images/},{images/Percorsi_fin/}}
\begin{document}
\begin{frontmatter}

\title{Interacting With a Mobile Robot with a Natural Infrastructure-Less Interface} 

\author{Valeria Villani,}
\author{Lorenzo Sabattini,}
\author{Giuseppe Riggio,}
\author{Alessio Levratti,} 
\author{Cristian Secchi,}
\author{Cesare Fantuzzi}

\address{Department of Sciences and Methods for Engineering (DISMI), University of Modena and Reggio Emilia, Reggio Emilia, Italy, \\ (e-mail name.surname@unimore.it)}

\begin{abstract}
% Abstract of not more than 250 words.
In this paper we introduce a novel approach that enables users to interact with a mobile robot in a natural manner. The proposed interaction system does not require any specific infrastructure or device, but relies on commonly utilized objects while leaving the user's hands free. Specifically, we propose to utilize a smartwatch (or a sensorized wristband) for recognizing the motion of the user's forearm. Measurements of accelerations and angular velocities are exploited to recognize user's gestures and define velocity commands for the robot.
The proposed interaction system is evaluated experimentally with different users controlling a mobile robot and compared to the use of a remote control device for the teleoperation of robots. Results show that the usability and effectiveness of the proposed natural interaction system based on the use of a smartwatch provide significant improvement in the human-robot interaction experience.
\end{abstract}


\end{frontmatter}
%===============================================================================

\section{Introduction}
%\vv{In recent decades the development of highly advanced robotic systems has seen them spread throughout our daily lives in several application fields, such as social assistive robots [1, 2], surveillance robots [3] and tour-guide robots [21], etc. As a consequence, the development of new interfaces for human-robot interaction (HRI) has received increasing attention in order to provide a more comfortable means of interacting with remote robots and encourage non-experts to interact with robots.}

In this paper we propose a novel methodology for letting a user interact with a mobile robot in a \emph{natural} manner. While traditional robotic systems are typically utilized in industrial contexts by specialized users, recent advances in robot control and safety systems are extending the field of application to diversified domains, including social assistance~\citep{bemelmans2012}, surveillance~\citep{dipaola2010}, tour-guide~\citep{tomatis2002} or floor cleaning~\citep{kang2014}. Furthermore, in recent years costs related to the installation of robotic equipments have been dramatically decreasing, thus allowing also small industrial places to invest in such systems. As a consequence, robot users are significantly increasing, both in number and in variety, and inexperienced users have often the necessity of utilizing robotic systems. 

A very important application in which mobile robots have been increasingly used in the last few years is goods transportation in industrial environments. The most well known examples are represented by Automated Guided Vehicles (AGVs) used for logistic operations~\citep{diganitase2015}: %,wurman2008}:
 these systems are fully autonomous, and the intervention of the user is limited to supervision operations. However, mobile robots can be also utilized as users' assistants, for goods transportation, in small-scale industrial scenarios. An example is represented by the system described in~\citep{Levratti_2016}: in this application, the mobile robot assists the operators of tire workshops, for transporting heavy loads (in this case, the wheels). This is a typical example in which, generally, operators do not have any specific robotic-related education: it is then mandatory to equip the robots with effective user interfaces, for the successful application of robotic systems in those contexts. In the last few years,  the concept of natural user interfaces (NUIs) has been formalized. This paradigm entails utilizing direct expression of mental concepts by intuitively mimicking real-world behavior. NUIs provide a natural way, for the user, to interact with robotic systems, since the interaction is reality-based and exploits users' pre-existing knowledge, utilizing actions that correspond to daily practices in the physical world \citep{Jacob_2008}. This is achieved, for instance, by letting users directly manipulate objects, thus spatially and physically interacting with robots, rather than programming them by typing commands on a keyboard (which is the traditional programming paradigm). Novel technologies are then utilized, such as voice, gestures, touch and motion tracking \citep{Blake_2011, Hornecker_2006}.%\citep{Bandeira_2015, Blake_2011, Hornecker_2006}.

In the literature, one of the most frequently utilized paradigms for providing intuitive interaction is gesture recognition. Gestures are typically recognized by means of colored markers~\citep{Raheja_2012}, or directly detecting human palm~\citep{Kaura_2013}. Most of the techniques available in the literature exploit vision systems~\citep{Sanna_2013, DeLuca_2012, Raheja_2012, Kaura_2013}. %\citep{Waldherr_2000, Sigalas_2010}.
 One of the main drawback of such solutions is that, typically, vision systems require a careful control of the environmental conditions, in terms of proper lighting conditions and camera angles~\citep{Raheja_2012}. These conditions are often too restrictive, in real application scenarios, in particular in outdoor applications. 

Moreover, it is worth noting that these interaction systems require a \emph{dedicated infrastructure}: this implies that these systems are not portable, since they are effective only when the user lies in the field of view of the sensors, that is in front of the camera~\citep{Sanna_2013}. Furthermore, these systems cannot track the robot since the user can barely move and follow it.

The method proposed in~\citep{Levratti_2016} partially overcomes the latter problem: in this application, the camera is mounted on the robot, and the user is allowed to follow the robot in its movements. However, also in this case, the user must stand in front of the camera in order for the gestures to be recognized. While gestures are utilized for providing the robot with high level commands, they are not suitable for precisely controlling the robot's trajectory. For this purpose, in~\citep{Levratti_2016} the authors introduce the use of a remote control device (the Geomagic Touch) for the teleoperation of the robot. Although such a device provides the user with a haptic feedback, the main drawback of this interaction system is in the fact that it lacks embeddedness in real space \citep{Hornecker_2006}, since the user is forced to sit at a desk, to teleoperate the robot. Instead, exploiting intuitive human spatial skills by letting the user move in real space while interacting and providing good spatial mappings between objects and the task are essential \citep{Sharlin_2004}. Thus, such interaction system appears to be mainly suited for teleoperated applications, where the distance between the user and the robot is intrinsic to the application, and is not introduced solely by the human-robot interface.

In~\citep{Villani_2017_RAL} we proposed a novel interaction system that overcomes the aforementioned main drawbacks of traditional systems. Specifically, we proposed  a novel hands-free infrastructure-less natural interface for human-robot interaction, based on wrist-motion recognition. We utilized the term \emph{infrastructure-less} for indicating the fact that the proposed interaction system does not require any additional dedicated equipments (such as a camera, for example), thus enforcing portability, without causing any limitation in the  physical area where interaction occurs. The user has complete freedom of motion, which makes the interaction very natural and intuitive.
By means of the proposed interaction system, the user has freedom of movement and is immersed in the environment where the robot moves, being able to track it with non-fragmented visibility \citep{Hornecker_2006}. This greatly improves the usability and effectiveness of interaction, in particular in those scenarios where the user shares the same working area as the robot, since she/he can control the robot standing in its proximity.

Specifically, the proposed interaction system is based on the use of an everyday device that allows recognition of the wrist and arm motion: in particular, without loss of generality, we utilized a commercial smartwatch, which could be easily replaced by a  common  wristband for activity tracking. In general, any commercial device equipped with accelerometers, gyroscopes and/or magnetometers, which can therefore measure movements of the wrist of the user, fits the proposed approach.
In~\citep{Villani_2017_RAL} we considered, as a target application, a quadrotor used for inspection operations. In this paper, we extend the proposed methodology to the case of a mobile robot, which can be used for heavy loads transportation in industrial and domestic applications. As a basis for comparison, we will consider the system recently presented in~\citep{Levratti_2016}.

\section{Multi-modal control architecture for the interaction with a  mobile robot}\label{sec:mobilerobot}

\subsection{Overview of the system}\label{subsec:overview}

\begin{figure}%[tbh]
	\centering
	{\includegraphics[width=\columnwidth]{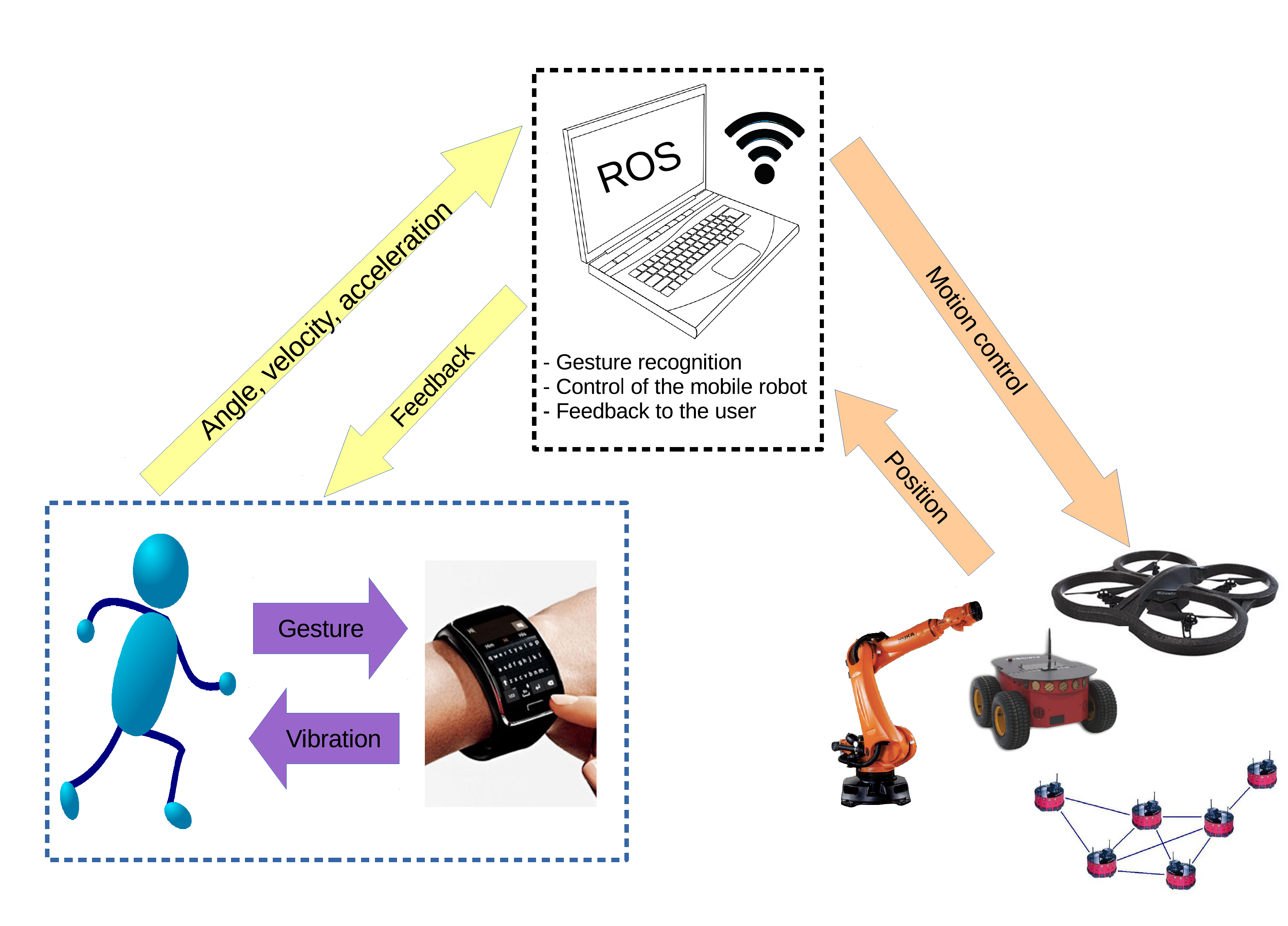}}
	\caption{Overview of the proposed interaction and control system.}%
	\label{fig:overview_general}%
\end{figure}

In this paper we extend the interaction and control system first introduced in~\citep{Villani_2017_RAL} and depicted in Fig.~\ref{fig:overview_general}. The system is composed of different interconnected elements: the user wears the smartwatch or the activity tracker wristband, which is used to acquire the motion of her/his arm and wrist. Specifically, characteristic measurements related to the motion are considered, such as angular displacement, angular velocity and linear acceleration. 
A mobile robot is considered, and, in general, it is supposed to be equipped with a motion control and a localization system: since these problems have been widely addressed in the literature, they will not be considered in this paper (see, for instance, \citep{campion2008} and references therein).

A data processing system is then implemented (on the smartwatch, on the robot on-board control system, or on a separate computation unit) for translating the user's motion into commands for the robot. Specifically, the motion of the user is analysed for recognizing gestures (used for imposing high-level commands to the robot), or for defining velocity commands (i.e. the robot's velocity is controlled as a function of the user's motion).

Feedback can be provided to the user, in terms of modulated vibration frequency of the smartwatch that informs her/him about the current status of the system.

\subsection{Control architecture for a wheeled mobile robot}\label{subsec:control_architecture}

In the following we describe a specific realization of the proposed interaction system: in particular, we consider a smartwatch utilized for letting a human operator interact with a wheeled ground mobile robot. 

%\ls{Brief description of the differential drive kinematics}

We will hereafter consider a wheeled mobile robot, moving in a two-dimensional space. The configuration of the mobile robot is defined by three variables $\left[x,y,\theta\right]$. In particular, $\left[x,y\right]\in\mathbb{R}^2$ represent the position of a representative point of the robot (e.g. its barycenter, the center of the wheel axis, etc.) with respect to a global reference frame, and $\theta\in \left[0, 2\pi\right)$ represents the orientation of the robot.
In this paper we will consider a mobile robot with differential drive kinematics, namely a mobile robot with two independently actuated wheels. This choice is motivated by the fact that this kind of robot is quite common in several applications, and its simple kinematic structure allows to keep the notation simple. Nevertheless, the proposed methodology can be trivially extended to consider more complex kinematic and dynamic models. For further details, the reader is referred to~\citep{campion2008}.

Define now $\omega_R,\omega_L \in \mathbb{R}$ as the angular velocities of the right and left wheels, respectively. Moreover, let $r>0$ be the wheel radius, and $d>0$ be the distance between the two wheels. Hence, it is possible to define the linear and angular velocities of the robot, $v$ and $\omega$, respectively, as follows:
\begin{equation}
v = \dfrac{r\left(\omega_R + \omega_L\right)}{2}, \quad
%\end{equation} 
%\begin{equation}
\omega = \dfrac{r\left(\omega_R - \omega_L\right)}{d}.
\label{eq:vel}
\end{equation} 
Subsequently, the kinematic model of the differential drive robot can be written as follows:
\begin{equation}
\left[\begin{array}{c}
\dot{x}\\ \dot{y} \\ \dot{\theta}
\end{array}\right] = \left[\begin{array}{c}
\cos \theta\\ \sin \theta \\ 0
\end{array}\right]v + 
\left[\begin{array}{c}
0 \\ 0 \\ 1
\end{array}\right]\omega.
\label{eq:kin}
\end{equation}

%\vspace{.7cm}
%\ls{Gesture recognition}

The interaction and control architecture introduced in Subsection~\ref{subsec:overview} requires the system to identify \emph{gestures} performed by the user: these gestures are then utilized for changing the operational mode of the robot. A methodology for recognizing gestures from the arm and wrist motion was introduced in~\citep{Villani_2017_RAL} and \citep{Villani_2016_HMS_smartwacth}: this method is based on template matching \citep{Duda_1973}.%, Gonzalez_2008}.
This classical approach represents a standard algorithm for this kind of applications and was shown to exhibit good performance. 

Specifically, data acquired by the accelerometer, gyroscope and magnetometer of the smartwatch are utilized. A template is built, off-line, for each desired gesture: subsequently, the user's motion is compared, at run time, with those templates, by using template matching. A decision whether a gesture has just occurred is then taken, on-line, by computing suited metrics of comparison.
In particular, training data consisting of 60 repetitions of each wrist gesture are used to build templates. Signal epochs containing each repetition are manually selected and, then, aligned based on the normalized cross-correlation function \citep{Proakis_1996}. Thus, templates are built by averaging aligned epochs. As a metric of comparison for template matching we use the correlation coefficient \citep{Kobayashi_2012}. Further details and the experimental validation of this methodology for gesture recognition are in~\citep{Villani_2017_RAL}.% and \citep{Villani_2016_HMS_smartwacth}.

%\ls{Gestures can be used for changing the status. Short: how to find gestures. We use only the circumference}

While this approach can be used, in principle, with any set of gestures, we defined templates, without loss of generality, for the following gestures:
\begin{enumerate}
	\item \emph{Up}: sharp movement upwards,
	\item \emph{Down}: sharp movement downwards,
	\item \emph{Circle}: movement in a circular clockwise shape,
	\item \emph{Left}: sharp movement to the left,
	\item \emph{Right}: sharp movement to the right.
\end{enumerate}
These gestures represent a common choice~\citep{liu2009}, and have been chosen since they represent movements that are easy to reproduce, without requiring previous training. At the same time, these movements are not likely to happen accidentally, as the user moves in her/his daily activities.

%\ls{Description of the sad state machine}

\begin{figure}%[tbh]
	\centering
	{\includegraphics[width=0.65\columnwidth]{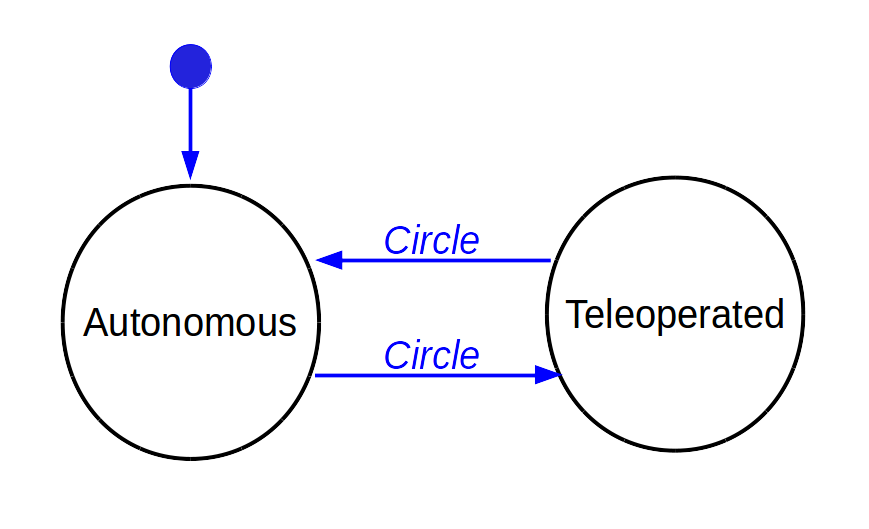}}
	\caption{State machine of the control algorithm.}
	\label{fig:state_machine}
\end{figure}

The high level control architecture utilized for controlling the mobile robot is described by the state machine diagram depicted in Fig.~\ref{fig:state_machine}.

The system is initialized in the \emph{Autonomous} state. In this state there is no interaction between the user and the mobile robot. Hence, the mobile robot is controlled by means of the internal, preprogrammed, control algorithm. If no control algorithm is defined, as in the application presented in this paper, then the mobile robot is stopped. 

Using the \emph{Circle} gesture, the user can move the system to the \emph{Teleoperated} state. In the \emph{Teleoperated} state, the user can directly control the motion of the robot. It is possible to go back to the \emph{Autonomous} state using again the \emph{Circle} gesture. 
It is worthwhile noting that, among the gestures we are able to detect, the \emph{Circle} is the only one exploited in the current implementation of the system. However, the state machine in Fig.~\ref{fig:state_machine} can be easily extended to include the other gestures, which could be utilized, as an example, to implement semi-autonomous behaviors such as tracking predefined trajectories, or following the user.%{to let the robot follow predefined trajectories or command it to follow the user.}

%\ls{Teleoperation}

\begin{figure}%[tbh]
	\centering
	{\includegraphics[width=0.45\columnwidth]{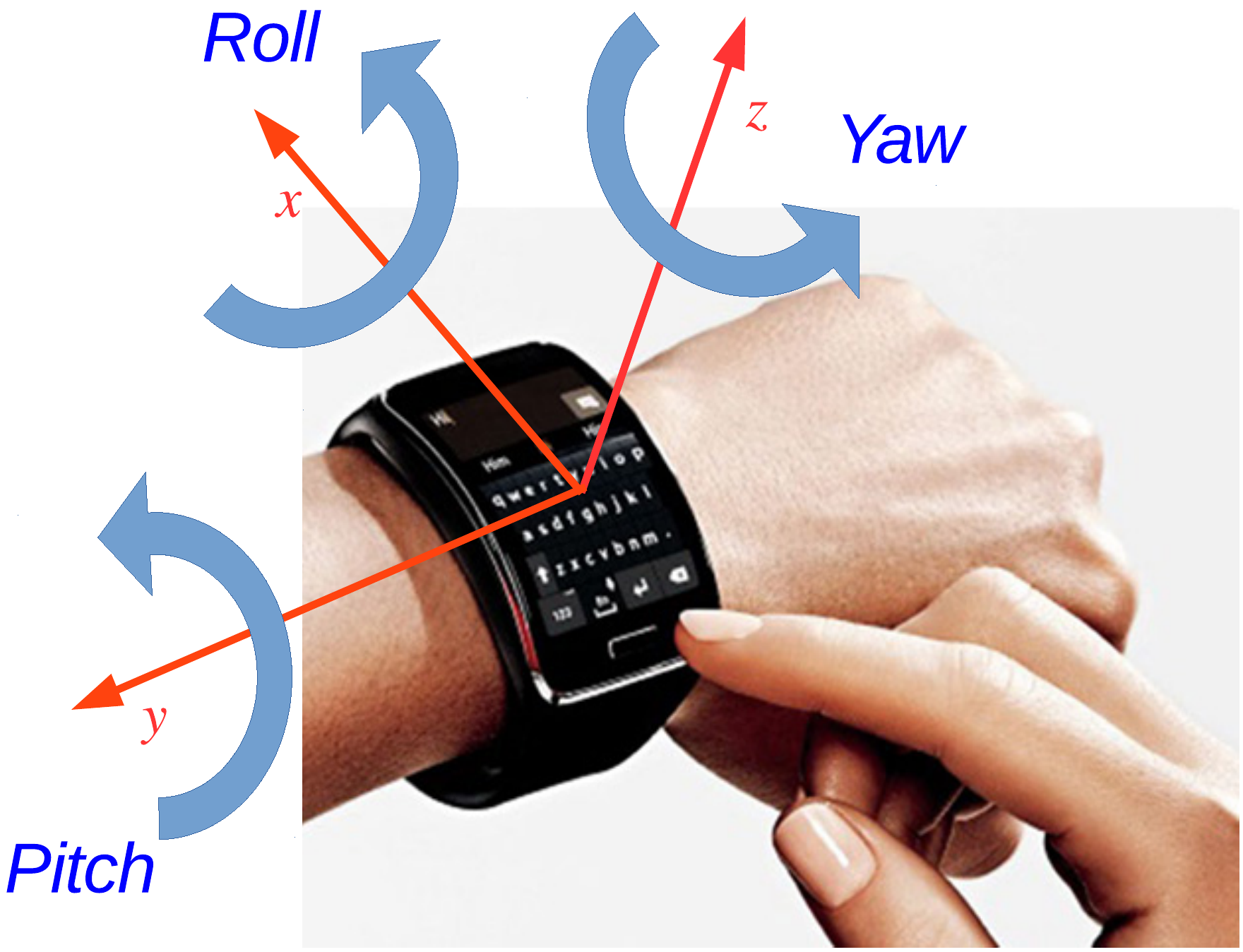}}
	\caption{Angles of the smartwatch.}%
	\label{fig:smartwatch_angles}%
\end{figure}

In the \emph{Teleoperated} state, the pose of the smartwatch is translated into a velocity command for the robot: the angles of the smartwatch are directly translated into velocity control inputs for the robot, in a natural and intuitive manner. 
To this end, refer to Fig.~\ref{fig:smartwatch_angles} for the definition of the smartwatch angles, and consider the kinematic model of the robot introduced in~\eqref{eq:kin}. The \emph{Roll} angle is used to control the linear velocity of the mobile robot. In particular, let $\vartheta_r \in \left[-\pi/2, \pi/2 \right]$ be the roll angle. Then, the linear velocity command is computed as follows:
\begin{equation}
v = K_r \, \vartheta_r
\label{eq:linearvelocity}
\end{equation} 
where $K_r > 0$ is a constant defined in such a way that the maximum angle that is achievable with the motion of the wrist corresponds to the maximum linear velocity of the mobile robot.

In a similar manner, the \emph{Pitch} angle is used to control the angular velocity of the mobile robot. In particular, let $\vartheta_p \in \left[-\pi/2, \pi/2 \right]$ be the pitch angle. Then, the angular velocity command is computed as follows:
\begin{equation}
\omega = K_p \, \vartheta_p
\label{eq:angularvelocity}
\end{equation} 
where the constant $K_p > 0$ is defined similarly to $K_r$ in~\eqref{eq:linearvelocity}.

%We would like to remark that, even though, in the proposed application, we utilize only the \emph{Circle} gesture, the remaining gestures can be exploited for implementing advanced featured and for considering additional behaviors for the robot.

\subsection{Feedback to the user}

The vibration of the smartwatch is utilized for providing the user with a haptic feedback. In particular, in this application we utilize vibration to acknowledge the user's command, that is for informing  the user about recognition of gestures. In particular, a single short vibration notifies the user when a gesture has been recognized by the control architecture. This is very useful for providing the user with an immediate feedback about the current status of the robot with reference to Fig.~\ref{fig:state_machine}, and then on the possibility of teleoperating it. This significantly increases situation awareness.

It is worth noting that such a feedback aims at solving, at least partially, the ephemerality of gestures. As discussed in~\citep{Norman_2010}, the ephemeral nature of gestures implies that, if a gesture is not recognized or is misrecognized, the user has little information available to understand what happened. This kind of sensory feedback tackles the first condition, by informing the user whether a gesture has been recognized. Misrecognition has not been considered yet, in the current implementation: this can be solved utilizing visual information on the screen of the smartwatch, or speech synthesis methods. However, as shown in~\citep{Villani_2017_RAL}, we experimentally verified that gesture misrecognition rate is quite low, which significantly reduces the need for this additional information.

\section{Experimental setup}\label{sec:experimental_setup}

To experimentally validate the proposed HRI system, we implemented the interaction and control system proposed so far. Specifically, the proposed system was implemented utilizing a Pioneer P3-AT mobile robot and a Samsung Gear S smartwatch. 
Due to the limited computation capabilities of the elaboration unit utilized for the control of the mobile robot, it was not possible to implement the gesture recognition and control architecture on-board the robot itself. Therefore, an external computer was utilized, and the overall architecture was implemented in ROS~\citep{quigley2009}.

%\begin{figure}
%\centering
%{\includegraphics[width=0.9\columnwidth]{images/overview_pioneer}}
%\caption{Implementation of the proposed interaction and control system.}
%\label{fig:experimental_setup}
%\end{figure}

The experimental setup is then implemented as follows. 
\begin{itemize}
	\item The user wears the smartwatch on her/his right wrist: the motion of her/his forearm is then acquired by the smartwatch.
	\item Characteristic measurements related to the motion (namely accelerations and angular velocities) are then sent from the smartwatch to an external computer, via Wi-Fi communication.
	\item The computer is in charge of processing the received data, to perform gesture recognition and for computing the velocity commands, as defined in~\eqref{eq:linearvelocity} and~\eqref{eq:angularvelocity}.
	\item Commands are then forwarded to the mobile robot via Wi-Fi communication.
	\item An acknowledgement is sent, via Wi-Fi, from the computer to the smartwatch, as soon as a gesture has been recognized. A short vibration is subsequently imposed to the smartwatch.
\end{itemize}

\section{Evaluation results}\label{sec:results}

Experiments aimed at evaluating the usability of the proposed interaction mode with the mobile robot. Additionally to the features of being hands-free and infrastructure-less, the objective of this evaluation was to assess the real benefit of the embeddedness in real space enabled by the use of the smartwatch.
For this purpose, the proposed interaction system was compared to a simple remote control system implementing unilateral teleoperation. For this purpose, we used the Geomagic Touch device: the user can move the end-effector of the device, and the motion is translated into a velocity command for the mobile robot. 
%For the sake of brevity, hereafter we mention the proposed interaction system as \emph{smartwatch} and the remote control system implementing unilateral teleoperation as \emph{remote control device}.

%\begin{figure}
%	\centering
%	{\includegraphics[width=0.6\columnwidth]{images/geomagic-touch}}
%	\caption{Geomagic Touch device.}
%	\label{fig:geomagic}
%\end{figure}

Three different experiments were implemented, involving 13 users (in each experiment), for assessing the usability and the effectiveness of the proposed strategy. All the experiments were implemented both with the smartwatch and the remote control device. Details are given in the following; a video showing the experiments can be found at http://tinyurl.com/smartwatchVillani-IFAC2017.

\subsection{Driving the robot through a cluttered environment}
%\ls{slalom tra i birilli}

\begin{figure}
	\centering
	{\includegraphics[width=.9\columnwidth]{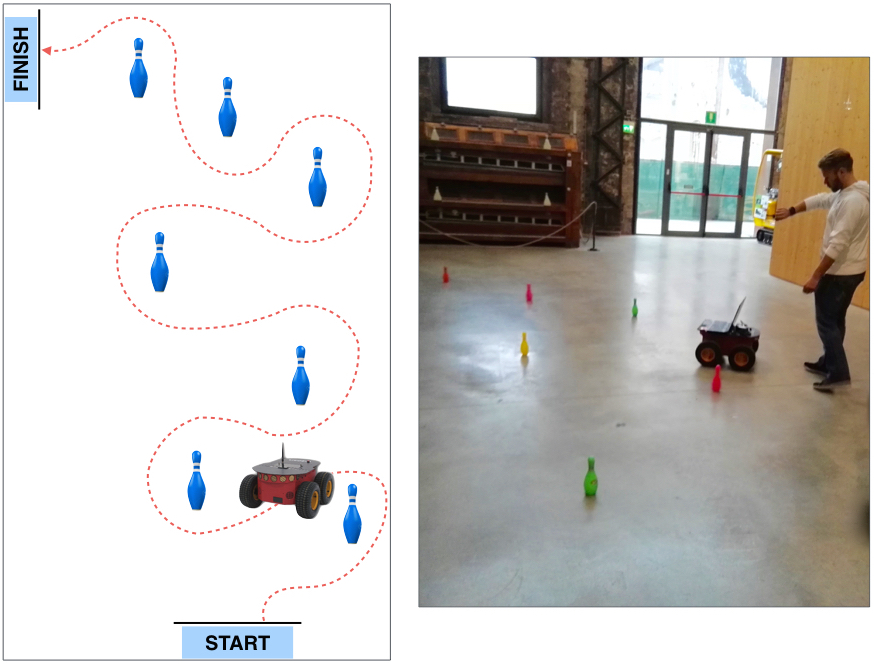}}	
	\caption{Driving through a cluttered environment: experimental setup.}
	\label{fig:slalom}
\end{figure}

In this experiment, the user was requested to drive the robot through a cluttered environment. In particular, the user had to move the robot in an area of $3.3 \times 9.0 \,[m^2]$, where seven plastic pins were placed on the ground, as shown in Fig.~\ref{fig:slalom}. The user was instructed to drive the robots from the initial position to the goal position, performing a \emph{slalom} path, without touching any pin. Each user was requested to perform the experiment with both the interaction systems.

For comparison purposes, we measured the total travel time, considering a penalty of $5\,[s]$ for each touched pin (if any). The results are summarized in Table~\ref{tab:slalom_results}.
The results achieved with the smartwatch are significantly better than the ones obtained with the remote control device ($p<10^{-3}$): while pins are typically avoided with both interaction systems, the smartwatch leads to a reduction of the execution time of approximately $33$\%.

\begin{table}
	\caption{\label{tab:slalom_results} Driving through a cluttered environment: results.}
	\centering
\begin{tabular}{ | c | c | c | c | c | c | c | }
	\hline
	 & \multicolumn{3}{|c|}{\textbf{Smartwatch}}     & \multicolumn{3}{|c|}{\textbf{Remote control}}  \\ \hline
	User & \pbox{1cm}{Travel time $[s]$} & Pins & \pbox{1cm}{Total time $[s]$} & \pbox{1cm}{Travel time $[s]$} & Pins & \pbox{1cm}{Total time $[s]$} \\ \hline
	1 & 58 & 0 & 58 & 107 & 1 & 112 \\ \hline
	2 & 57 & 0 & 57 & 93 & 0 & 93 \\ \hline
	3 & 65 & 0 & 65 & 82 & 0 & 82 \\ \hline
	4 & 58 & 0 & 58 & 88 & 0 & 88 \\ \hline
	5 & 66 & 0 & 66 & 76 & 0 & 76 \\ \hline
	6 & 78 & 0 & 78 & 92 & 1 & 97 \\ \hline
	7 & 106 & 0 & 106 & 97 & 0 & 97 \\ \hline
	8 & 80 & 0 & 80 & 129 & 2 & 139 \\ \hline
	9 & 60 & 0 & 60 & 90 & 0 & 90 \\ \hline
	10 & 68 & 0 & 68 & 100 & 0 & 100 \\ \hline
	11 & 80 & 1 & 85 & 200 & 0 & 200 \\ \hline
	12 & 79 & 0 & 79 & 115 & 0 & 115 \\ \hline
	13 & 61 & 0 & 61 & 97 & 0 & 97 \\ \hline \hline
	\textbf{Mean}  & \textbf{70.46} & & \textbf{70.85} & \textbf{105.08} & & \textbf{106.62} \\ \hline
\end{tabular}
\end{table}

\subsection{Visiting a set of targets}
%\ls{abbattere i birilli}

\begin{figure}
	\centering
	{\includegraphics[width=.9\columnwidth]{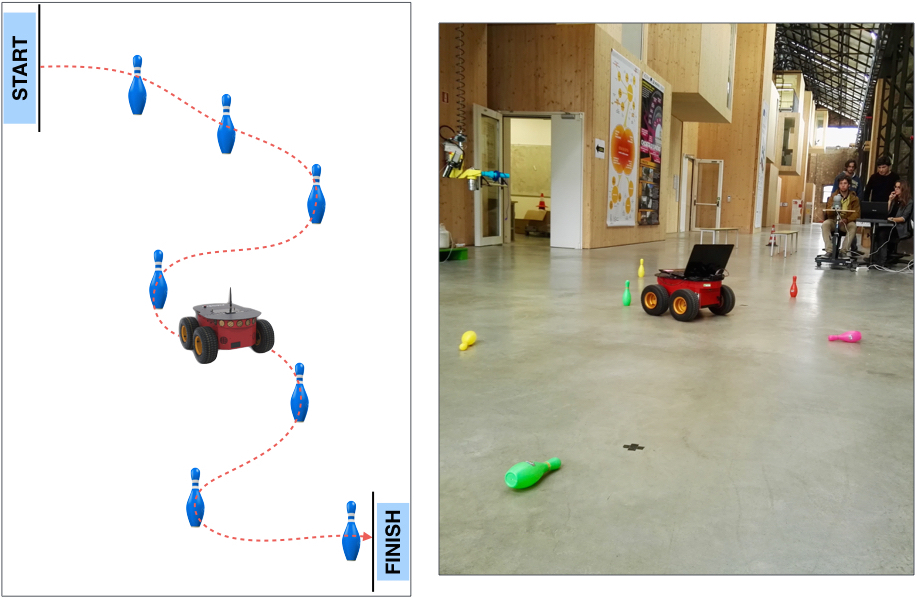}}
	\caption{Visiting a set of targets: experimental setup. %\ls{EVIDENZIARE IL PERCORSO E LA POSIZIONE INIZALE}
		}
	\label{fig:targets}
\end{figure}

In this experiment, the user was requested to drive the robot to visit a sequence of seven target locations, identified by plastic pins placed on the ground, as depicted in Fig.~\ref{fig:targets}. Each target position was considered as visited as soon as the corresponding pin was knocked down. %\footnote{\ls{Non sapevo come dire "abbattere i birilli". Ho quindi cercato su google "objective of bowling", ed \'e saltato fuori questo}}
 Each user was requested to perform the experiment with both the interaction systems. For comparison purposes, we measured the total time needed for knocking down all the pins.

The results are summarized in Table~\ref{tab:target_results}. The results achieved with the smartwatch are statistically significantly better than the ones obtained with the remote control device ($p=0.002$): in fact, the smartwatch leads to a reduction of the execution time of approximately $45$\%.

\begin{table}
	\caption{\label{tab:target_results} Visiting a set of targets: results.}
	\centering
	\begin{tabular}{ |  c | c | c | }
		\hline
		& {\textbf{Smartwatch}}   & {\textbf{Remote control}}  \\ \hline
		User & {Total time $[s]$} &  {Total time $[s]$} \\ \hline
		1 & 42 & 135  \\ \hline
		2 & 35 & 37  \\ \hline		
		3 & 28 & 46   \\ \hline		
		4 & 27 & 32  \\ \hline		
		5 & 40 & 45 \\ \hline
		6 & 38 & 53 \\ \hline
		7 & 47 & 64 \\ \hline
		8 & 39 & 110 \\ \hline
		9 & 31 & 52 \\ \hline
		10 & 36 & 67 \\ \hline
		11 & 53 & 78 \\ \hline
		12 & 40 & 107 \\ \hline
		13 & 33 & 56 \\ \hline \hline
		\textbf{Mean} & \textbf{37.62} & \textbf{67.85} \\ \hline
	\end{tabular}
\end{table}

\subsection{Exploring a building}
%\ls{percorso}

\begin{figure}
	\centering
	{\includegraphics[width=0.848\columnwidth]{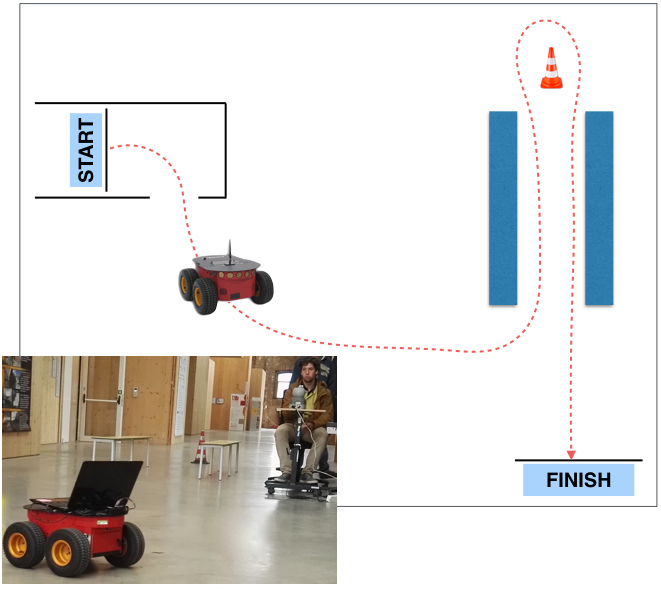}}
	\caption{Exploring a building: experimental setup.}
	\label{fig:building}
\end{figure}

In this experiment, the user was requested to drive the robot to perform a path that emulates the exploration of a building. The environment is depicted in Fig.~\ref{fig:building}. One of the advantages of the interaction system proposed in this paper is in the fact that the user can move along with the robot, while exploring an environment. Conversely, utilizing the considered remote control device, the user is forced to sit at a desk: in this case, live streaming of images acquired from a camera installed on-board the mobile robot was provided to the user, to help her/him to drive the robot.

The results are summarized in Table~\ref{tab:building_results}. The results achieved with the smartwatch are statistically significantly better than the ones obtained with the other device \mbox{($p\simeq10^{-7}$)}: in fact, the smartwatch leads to a reduction of the execution time of approximately $55$\%.
The results of this experiment show that the use of the remote control device requires a big effort to the user to control the robot standing in a fixed position (i.e., sitting at the desk where the device is located) and not relying on direct view. On the contrary, being able to follow the robot and sharing the same physical space by means of the proposed interaction methodology increase the effectiveness of the interaction, since human spatiality is exploited \citep{Sharlin_2004}.
%	It is worth noting that in this experiment the use of the remote control device highlighted the effort required to the user to control the robot standing in a fixed position, whereas when u 
%	the use of the smartwatch returned the greatest improvement over the remote control device

\begin{table}
	\caption{\label{tab:building_results} Exploring a building: results.}
	\centering
	\begin{tabular}{ |  c | c | c | }
		\hline
		& {\textbf{Smartwatch}}     & {\textbf{Remote control}}  \\ \hline
		User & {Total time $[s]$} &  {Total time $[s]$} \\ \hline
		1 & 44 & 105  \\ \hline
		2 & 43 & 71  \\ \hline		
		3 & 42 & 133   \\ \hline		
		4 & 43 & 83  \\ \hline		
		5 & 47 & 109 \\ \hline
		6 & 51 & 100 \\ \hline
		7 & 46 & 117 \\ \hline
		8 & 51 & 82 \\ \hline
		9 & 42 & 96 \\ \hline
		10 & 46 & 102 \\ \hline
		11 & 57 & 131 \\ \hline
		12 & 57 & 95 \\ \hline
		13 & 49 & 145 \\ \hline \hline
		\textbf{Mean} & \textbf{47.54} & \textbf{105.31} \\ \hline
	\end{tabular}
\end{table}

\section{Conclusions}\label{sec:conclusions}

In this paper we introduced a novel HRI approach that allows a hands-free natural interaction with a mobile robot. In particular, the motion of the user's wrist is recognized by means of a general purpose device, such as a smartwatch, and exploited to recognize gestures and define control inputs for the robot. Haptic feedback is provided to the user by means of a modulated vibration that informs her/him about the recognition of gestures, thus increasing situation awareness.
The usability of the proposed HRI approach based on the smartwatch was experimentally evaluated and compared to the use of a remote control device for the teleoperation of the robot. Usability was measured as the time required to follow three  paths, having different goals and setups, by the two piloting modalities.
The use of the smartwatch proved to be more intuitive and easy, allowing, in the 97\% of the performed trials, to complete the tasks in much less than the time taken when using the haptic device.

As future works, we plan to improve the usability assessment, measuring the cognitive workload associated to the use of the smartwatch to interact with the robot. This information can be exploited to adapt the robot's behavior to the cognitive and emotional state of the user, in a scenario of affective robotics. %~\citep{Rani_2004}.
Moreover, we plan to extend the experimental validation of the proposed interaction approach by comparing it to a bilateral teleoperation system providing the user with haptic feedback for obstacle avoidance and target tracking.

\bibliography{biblio}                                  

\end{document}